\documentclass[10pt,twocolumn,letterpaper]{article}

\usepackage{cvpr}              
\definecolor{cvprblue}{rgb}{0.21,0.49,0.74}
\usepackage[pagebackref,breaklinks,colorlinks,allcolors=cvprblue]{hyperref}

\usepackage{algorithm}
\usepackage{algpseudocode}
\usepackage{booktabs,makecell,enumitem}
\usepackage{array}      
\usepackage{ragged2e}   
   
\usepackage{booktabs}    
\usepackage{tabularx}   
\usepackage{array}       
\usepackage{caption}
\usepackage{booktabs}
\usepackage{threeparttable}
\usepackage{graphicx}    
\usepackage{multirow}    
\usepackage{adjustbox}   
\usepackage{colortbl}  
\usepackage{subcaption}  
\usepackage{booktabs}    
\usepackage{caption}     
\usepackage{placeins} 

\def\paperID{} 
\def\confName{CVPR}
\def\confYear{2026}

\title{VideoThinker: Building Agentic VideoLLMs with LLM-Guided Tool Reasoning}
\author{Chenglin Li\textsuperscript{1,5},  Qianglong Chen\textsuperscript{1}, Feng Han\textsuperscript{2,5}, Yikun Wang\textsuperscript{2,5},Xingxi Yin\textsuperscript{1},Yan Gong\textsuperscript{1}, Ruilin Li\textsuperscript{3,5}, \\
, Yin Zhang\textsuperscript{1}\thanks{Corresponding Authors} , Jiaqi Wang\textsuperscript{4,5}\footnotemark[1] \\
  \textsuperscript{1}Zhejiang University, \textsuperscript{2}Fudan University, \textsuperscript{3}Wuhan University, \\
  \textsuperscript{4}Shanghai AI Lab, \textsuperscript{5}Shanghai Innovation Institute \\
}

\begin{document}
\maketitle
\begin{abstract}
Long-form video understanding remains a fundamental challenge for current Video Large Language Models (VideoLLMs). Most existing models rely on static reasoning over uniformly sampled frames, which weakens temporal localization and leads to substantial information loss in long videos. Agentic tools, such as temporal retrieval, spatial zoom, and temporal zoom, offer a natural way to overcome these limitations by enabling adaptive exploration of key moments. However, constructing agentic video-understanding data requires models that already possess strong long-video comprehension, creating a circular dependency.
We address this challenge with VideoThinker, an agentic VideoLLM trained entirely on synthetic tool-interaction trajectories. Our key idea is to convert videos into rich captions and employ a powerful agentic LLM to generate multi-step tool-use sequences in the caption space. These trajectories are subsequently grounded back to video by replacing captions with actual frames, yielding a large-scale interleaved video–tool reasoning dataset without requiring any long-form understanding from the underlying VideoLLM. 
Training on this synthetic agentic dataset equips VideoThinker with dynamic reasoning capabilities, adaptive temporal exploration, and multi-step tool use. Remarkably, VideoThinker significantly outperforms both caption-only LLM agents and strong VideoLLM baselines across long-video benchmarks, demonstrating the effectiveness of tool-augmented synthetic data and adaptive retrieval–zoom reasoning for long-form video understanding.
\end{abstract}

\begin{figure*}
    \centering
    \includegraphics[width=1.05\linewidth]{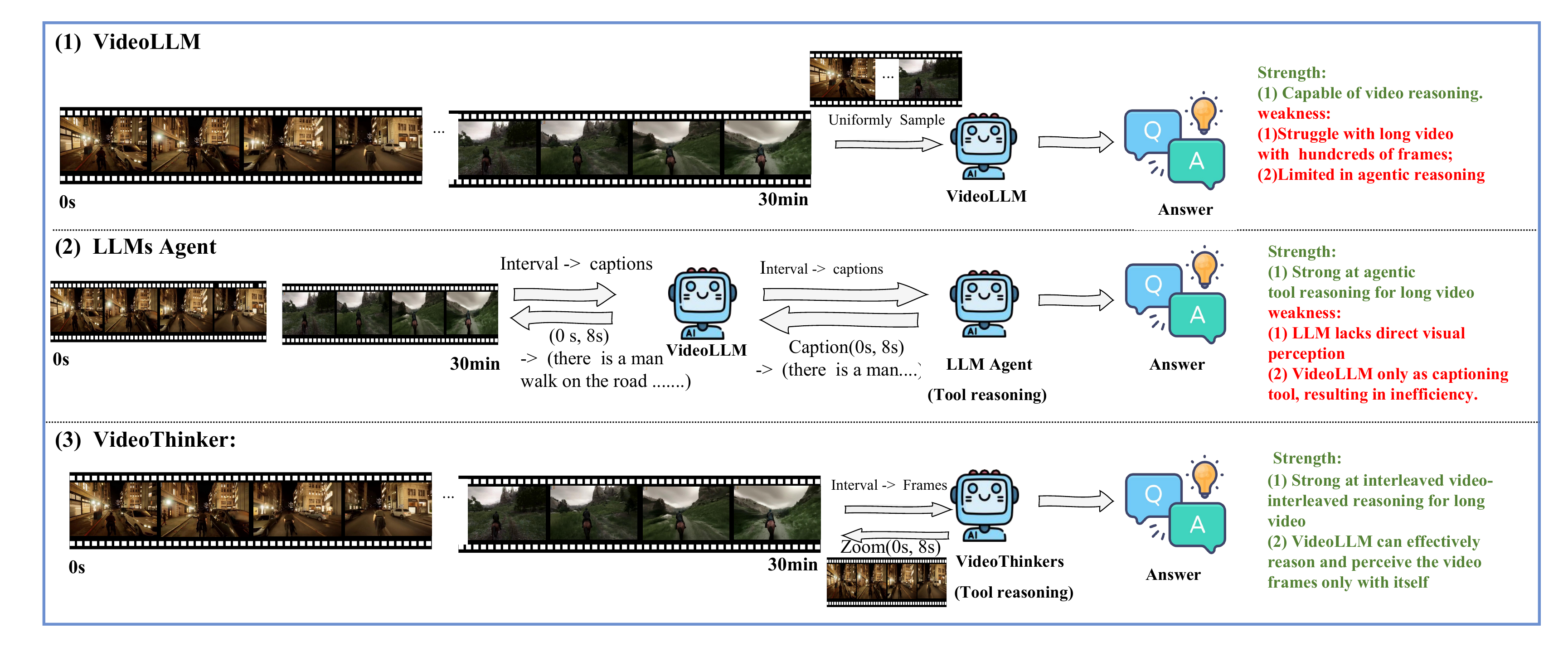}
    
\caption{Comparison of VideoThinker with VideoLLMs and LLM agents. VideoThinker excels at interleaved video reasoning on long videos, using agentic tools to iteratively perceive and reason over video frames step by step.}

    \label{fig:compare}
\end{figure*}

\section{Introduction}
Video understanding is a fundamental task that requires reasoning over both spatial and temporal dimensions of videos~\citep{yu2019activitynet,ning2023video,chen2023autoeval,li2024mvbench,li2024vcbench,zhou2025flattery,yang2026infact,li2026retrack,hu2026refine}. Recent advances in Video Large Language Models (VideoLLMs) have achieved remarkable progress in this area~\citep{li2023videochat,zhang2023video,lin2023video,li2024llava,bai2025qwen2,zhang2024video}. Despite these advances, existing models still struggle with long-form videos~\citep{wu2024longvideobench,fu2024video,wang2024lvbench}, where relevant visual evidence is sparse, distributed across time. Static reasoning over uniformly sampled frames often causes major information loss and poor temporal localization, while also being computationally costly for long videos with hundreds of frames.

A promising avenue for improving long-video understanding lies in the concept of interleaved multimodal reasoning, which involves combining visual perception with language reasoning in a dynamic and adaptive manner~\cite{zheng2025deepeyes,wang2025simple,lai2025mini}. Inspired by successful techniques in the image domain, such as those seen in OpenAI’s o3 model~\citep{openai_o3_o4mini_2025}, concurrent work attempts to extend such interleaved reasoning to videos, proposing the notion of \textit{thinking with videos}~\citep{meng2025openo3,zhang2025thinking,fu2025lover1}. However, these approaches face key limitations:
(1) they often rely on proprietary models such as Gemini-Pro~\citep{comanici2025gemini} to construct agentic video-understanding data, which requires extensive prompt engineering and multi-stage filtering; Moreover, although Gemini-Pro can produce interleaved video-reasoning traces, it still lacks the \textit{emergent} video-interleaved reasoning capability observed in o3's image reasoning~\citep{openai_o3_o4mini_2025}; and (2) they primarily perform single temporal zoom operations without a temporal retrieval mechanism, making them inefficient at identifying key temporal intervals. In contrast, open-source LLMs such as Qwen3~\citep{yang2025qwen3} and DeepSeek~\citep{liu2024deepseek} exhibit strong tool-augmented reasoning capabilities as language agents for long-form video understanding. They employ various tools, with VideoLLMs generating intermediate frame captions as tools to support video reasoning~\citep{wang2024videoagent,fan2024videoagent,yuan2025videodeepresearch,wang2025videotree}. However, in such frameworks, the VideoLLM serves merely as a passive captioning module, while the LLM itself cannot directly perceive visual information.

Motivated by these insights, we propose \textbf{VideoThinker}, an agentic VideoLLM trained entirely on synthetic tool-interaction trajectories. The core idea is to first convert videos into rich textual captions and then leverage a powerful agentic LLM to generate multi-step tool-use sequences within this caption space. To enable effective long-video reasoning, we design two complementary agentic tools: 
(1) \textbf{Temporal Retrieval}, which identifies candidate temporal intervals that may contain relevant information by leveraging audio transcripts (as subtitles), scene descriptions, and subtitle-based summaries; and 
(2) \textbf{Temporal Zoom}, which inspects intervals at a finer granularity through more detailed subtitles or frames. By leveraging these tools together with the LLM's tool-augmented reasoning capability, we construct multi-turn tool-interaction trajectories. When the LLM invokes the temporal-zoom tool, the VideoLLM generates intermediate textual captions, which are subsequently replaced with actual video frames to form fully video-interleaved reasoning data. These trajectories are then used to fine-tune the VideoLLM, enabling it to actively retrieve and perceive key frames during reasoning, effectively bridging the gap from text- and image-based reasoning to true video reasoning. Furthermore, VideoThinker incorporates a confidence-gated tool controller, achieving strong performance gains: +6.8\% on MLVU and +10.6\% on LVBench over vanilla VideoLLMs, and +3.9\% and +3.5\% over caption-only LLM agents equipped with our tools. Our main contributions are summarized as follows:

\begin{itemize}
\item We construct a high-quality agentic video–tool reasoning dataset guided by LLM-based tool reasoning. This synthetic data captures diverse temporal reasoning and tool-use behaviors essential for long video understanding.

\item We equip \textbf{VideoThinker} with dynamic reasoning, adaptive temporal exploration, and multi-step tool use via \textit{Temporal Retrieval} and \textit{Temporal Zoom} for efficient long-video understanding.

\item Extensive experiments demonstrate that \textbf{VideoThinker} outperforms strong VideoLLM and LLM-agent baselines across long-video benchmarks, validating the effectiveness of tool-augmented synthetic data and adaptive retrieval–zoom reasoning.
\end{itemize}

\section{Related Work}
\label{sec:formatting}

\begin{figure*}[t]
\centering
\includegraphics[width=1.0\linewidth]{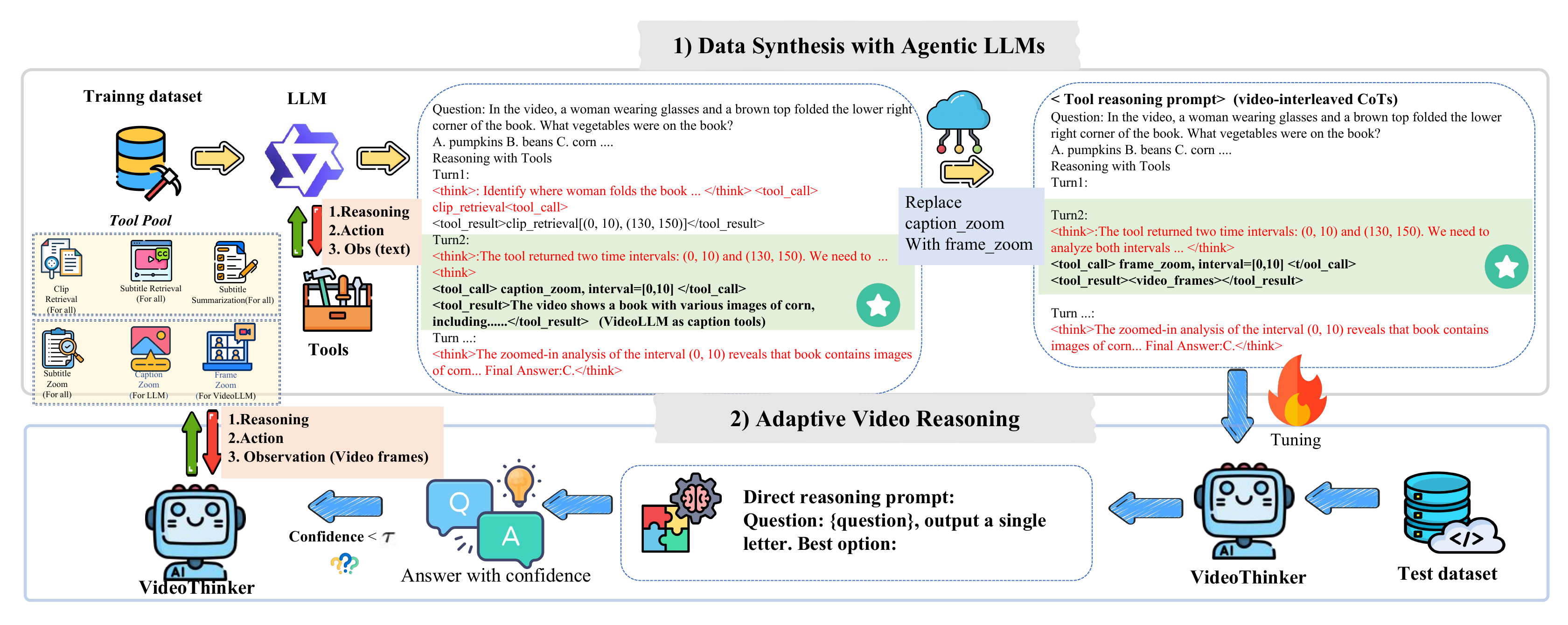}
\caption{VideoThinker integrates retrieval and zoom tools for multi-turn reasoning. LLMs use \textbf{caption\_zoom} to generate reasoning data from videos, which is later replaced by frame\_zoom and video frames to build agentic video understanding CoTs.}
\label{fig:framework}
\end{figure*}
\subsection{Agentic LLMs for Video Understanding}

Recent advances in large language models (LLMs), particularly their reasoning and planning capabilities, have spurred the development of agents~\cite{yao2022react,huang2024understanding}. By leveraging external tools~\cite{qin2023toolllm,yuan2024easytool}, LLMs can acquire information, plan, and execute actions in complex environments, bridging the gap between general-purpose language models and real-world applications. Extending this paradigm to video understanding, recent works integrate LLM reasoning with agentic tools to tackle complex video tasks without costly training. However, applying such agents to video understanding poses challenges, as many existing tools do not generalize well across temporal sequences. Several works have proposed video-specific adaptations of the LLM-agent paradigm~\cite{wang2024videoagent,fan2024videoagent,yuan2025videodeepresearch,li2025adaptive}. LifeLongMemory~\cite{wang2023lifelongmemory} constructs a text-based episodic memory from video narrations, enabling LLMs to reason and retrieve relevant information for downstream tasks. DoraemonGPT~\cite{yang2024doraemongpt} introduces a MCTS-based prompting strategy, combining tool use with structured memory for video reasoning. VideoAgent~\cite{wang2024videoagent} frames video understanding as a sequential decision-making process, where the agent decides whether to gather more information or produce an answer, mimicking human video interpretation strategies. VideoTree~\cite{wang2025videotree} enhances efficiency and interpretability by adaptively extracting key visual information in a coarse-to-fine manner and organizing it in a human-readable tree for LLM reasoning. Related approaches, such as DVD~\cite{zhang2025deep} and VideoExplorer~\cite{yuan2025thinkvideosagenticlongvideo}, employ agentic search strategies over segmented video clips to tackle long-form VideoQA. Together, these works demonstrate the potential of agentic LLMs for structured, tool-augmented reasoning in complex video understanding tasks. A key limitation of these approaches is that they rely on a strong LLM as the core reasoning engine, while VideoLLMs serve only as auxiliary tools (e.g., generating captions). In contrast, our method elevates the VideoLLM itself to the reasoning core, effectively simplifying the agent architecture. By combining the capabilities of a VideoLLM with the agentic planning paradigm, our framework achieves better performance and reduces reliance on LLMs, improves efficiency, and streamlines the integration of multimodal reasoning for long-form video understanding.

\subsection{Multimodal reasoning with MLLMs}
Multimodal reasoning differs from pure text reasoning in that it enables models to directly manipulate visual content, such as zooming into specific image regions~\cite{wang2025simple,lai2025mini,su2025openthinkimg,zheng2025deepeyes,wang2026towermind,zhao2024symmetric,di2025balancing,wang2025autoregressive}. Leveraging this capability, our work focuses on deciding which video clips to zoom in on, enabling efficient processing of long videos. Recent concurrent works, such as Open-o3-Video~\cite{meng2025openo3} and Video-MTR~\cite{xie2025video}, also adopt dynamic processing strategies using chain-of-thought reasoning (CoT) over video content. These approaches generate multimodal tool data for training advanced VideoLLMs with Gemini-2.5-pro. On one hand, Gemini-pro itself has not demonstrated “thinking with videos” capabilities via multimodal tools (contrary to O3, which shows “thinking with images” on image tasks). On the other hand, synthesizing high-quality multi-turn video reasoning data typically requires data validation and extensive data engineering, making direct generation costly. Differently, we construct the tool-augmented reasoning ability from the language agents and use video frames captions as textual proxies for multimodal tool outputs. This approach allows efficient data synthesis while leveraging open-source models such as Qwen3 to produce high-quality video-interleaved CoT. In this way, our framework efficiently equips VideoLLMs with the capability to “think with videos” through tool-augmented reasoning, without heavy computational or annotation costs.

\subsection{Long-form VideoQA and Video-LLMs}
Long-form VideoQA requires reasoning over extended temporal contexts and capturing causal dependencies across events~\citep{xiao2021next,wu2024longvideobench,fu2024video,wang2024lvbench,zhou2025flattery}. Recent Video-LLMs extend temporal encoders for joint spatio-temporal reasoning~\citep{bai2025qwen2,li2023videochat}, but still face memory and computation bottlenecks when processing long videos. To alleviate this, several methods convert videos into textual captions or keyframe summarization, allowing LLMs to perform text-based reasoning~\citep{wang2024videoagent,wang2025videotree}. While effective for scalability, these approaches lose fine-grained temporal details~\cite{zhang2025trimtokenator,zhang2025trimtokenatorlc}. Motivated by these limitations, we propose VideoThinker, which focuses on key segments and performs frame-level video-interleaved reasoning, enabling efficient long-form video understanding.

\section{Method}
In this section, we present VideoThinker, an agentic VideoLLM for long-form video understanding through dynamic, tool-augmented reasoning. To tackle temporal sparsity and information loss, we design two complementary tools—Temporal Retrieval and Temporal Zoom—that enable adaptive exploration of key moments. We then construct a synthetic training dataset based on tool-interaction trajectories from LLM, allowing the VideoLLM to learn multi-step reasoning and video-frame perception.


\begin{figure}[H]
    \centering
    \includegraphics[width=1.0\linewidth]{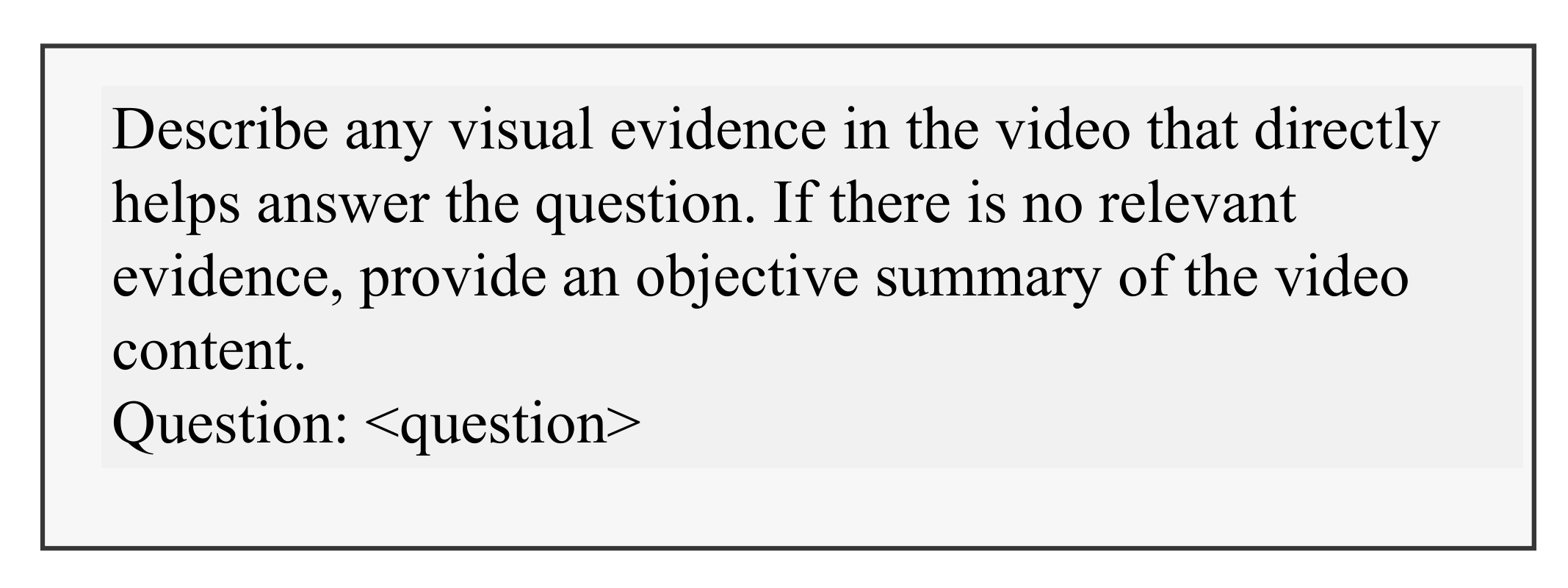}
    \caption{The prompt designed to enable VideoLLM to serve as a caption generation tool.}
    \label{figure:prompt_vllm}
\end{figure}

\subsection{Tools Design}
\label{method:1}
We design two complementary agentic tools for efficient retrieval and reasoning over long-form videos. It includes functional groups: Temporal Retrieval tools (Clip Retrieval, Subtitle Retrieval, Subtitle Summary) for multimodal content access, and Temporal Zoom tools (Frame Zoom, Subtitle Zoom) for fine-grained inspection of some intervals. This modular design enables adaptive focus on key segments while preserving global context. Under an agentic search paradigm, the agent decomposes queries, composes tool calls, and iteratively gathers evidence to refine understanding and localize relevant information within long videos.
\subsubsection{Temporal Retrieval}

\noindent\textbf{ClipRetrieval}  
This tool provides coarse-grained access to semantically relevant temporal regions in long videos. 
Given a \texttt{video\_path} and a text query, the video is first segmented into 10-second clips, each encoded using LanguageBind-Video~\cite{zhu2023languagebind} to obtain clip-level embeddings. 
The tool then retrieves the top-ranked clips with the highest semantic similarity to the input query and returns their corresponding temporal intervals. 
By iteratively invoking \texttt{ClipRetrieval} with refined queries informed by intermediate reasoning results, the agent can progressively zoom in on the most relevant video segments, efficiently narrowing the search space for subsequent analysis.
                                                                                                                                                                                                                                                                                                                                                                                                                                                                                                                                                                                                                                                                                                                                                                                                                                                                                                                                                                                                                                                                                                                                                                                                                                                                                                                                                                                                                                                                                                                                                                                                                                                                                                                                                                                                                                                                                                                                                                                                                                                                                                                                                                                                                                                                                                                                                                                                                                                                                                                                                                                                                                                                                                                                                                                                                                                                                                                                                                                                                                                                                                                                                                                                                                                                                                                                                                                                                                                                                                                                                                                                                                                                                                                                                                                                                                                                                                                                                                                                                                                                                                                                                                                                                                                                                                                                                                                                                                                                                                                                                                                                                                                                                                                                                                                                                                                                                                                                                                                                                                                                                                                             cvpr
\medskip
\noindent\textbf{SubtitleRetrieval}
This tool enables fine-grained text-level retrieval over automatically transcribed subtitles. 
Given a \texttt{video\_path} and a text query, Whisper~\citep{radford2023robust} is employed to transcribe the video audio stream, after which the tool retrieves subtitle segments most semantically relevant to the query, along with their timestamps. 
Through iterative refinement, \texttt{SubtitleRetrieval} allows the agent to align textual evidence with visual cues, thereby precisely localizing key temporal intervals.

\medskip
\noindent\textbf{SubtitleSummary}  
This tool generates concise, query-focused summaries to support global video understanding. 
Built upon Qwen3-30B~\citep{yang2025qwen3}, it processes the complete subtitle transcript to extract information most relevant to the input query. 
The resulting summaries enable the agent to efficiently grasp the overall narrative flow and contextual dependencies of long-form videos, serving as high-level semantic guidance for downstream reasoning.

\subsubsection{Temporal Zoom}
\noindent\textbf{FrameZoom}  
This tool supports fine-grained visual inspection by extracting raw frames within a specified temporal interval. 
Given start and end timestamps, it retrieves the corresponding frames for detailed analysis. For example, if a video contains \(32\) frames in total but only \(2\) fall within the interval \([0,10]\) seconds, invoking \texttt{FrameZoom(0,10)} resamples this interval to return \(8\) frames, thereby increasing visual density and improving perceptual detail for localized reasoning.

\medskip
\noindent\textbf{SubtitleZoom}  
This tool extracts subtitle segments corresponding to a specified temporal interval. By providing start and end timestamps, the agent obtains localized subtitle text aligned with the chosen segment, enabling fine-grained multimodal reasoning within temporally bounded contexts.

\medskip
\noindent\textbf{CaptionZoom}  
This tool acts as a semantic bridge between visual frames and text-based reasoning. 
It first invokes \texttt{FrameZoom} to extract frames from a given interval, then employs a VideoLLM to generate natural-language captions summarizing key visual events, objects, and interactions. 
During tool-reasoning data synthesis, the LLM agent queries \texttt{CaptionZoom} for localized visual semantics using textual descriptions rather than raw frames, effectively simulating perceptual grounding through tool-based interaction. 
In the final training stage, these generated captions are replaced by visual frame tokens, enabling the VideoLLM to internalize structured reasoning patterns grounded directly in visual representations.

\begin{algorithm}[t]
\caption{Tool-Reasoning Data Synthesis}
\label{alg:agentic_search_variant}
\begin{algorithmic}[1]
\Require Training sample \((v_i, x_i)\), max steps \(T\), LLM \(P\), VideoLLM, tool set \(\mathcal{M}\), action space \(\mathcal{A} = \mathcal{M} \cup \{\text{Answer}\}\)

\State Generate video caption: \(c_i \leftarrow \text{VideoLLM}(v_i)\)
\State Construct initial input prompt: \(p_i \leftarrow p(x_i, c_i)\)
\State Initialize reasoning history: \(\mathcal{H}_0 \leftarrow \{p_i, \mathcal{A}\}\)

\For{\(t = 1\) to \(T\)}
    \State Generate reasoning step: \(\hat{r}_t \leftarrow P.\text{reason}(\mathcal{H}_{t-1})\)
    \State Select action: \((\alpha_t, \hat{\rho}_t) \leftarrow P.\text{call}(\hat{r}_t, \mathcal{H}_{t-1})\)
    \If{\(\alpha_t = \text{Answer}\)}
        \State \textbf{break}
    \EndIf
    \State Invoke tool: \(\hat{\omega}_t \leftarrow \alpha_t(\hat{\rho}_t)\) 
    \State Update history: \(\mathcal{H}_t \leftarrow \mathcal{H}_{t-1} \cup \{(\hat{r}_t, \alpha_t, \hat{\omega}_t)\}\)
    \If{\(t = T\)}
        \State Produce final answer: \(\hat{y}_i \leftarrow P.\text{answer}(\mathcal{H}_t)\)
    \EndIf
\EndFor

\State \Return \(\hat{y}_i\), \(\mathcal{H}\)
\end{algorithmic}
\end{algorithm}

\subsection{Data Synthesis with Agentic LLMs}
\label{method:2}
Given a training dataset \(\mathcal{D} = \{(v_i, x_i, y_i)\}_{i=1}^M\), where \(v_i\) denotes a video, \(x_i\) is the corresponding question, and \(y_i\) is the ground-truth answer, we construct structured tool system prompts to elicit tool-based reasoning from the LLM. For each query \(x_i\), a video caption \(c_i\) of \(v_i\) is generated using the VideoLLM. The query \(x_i\), caption \(c_i\), and tool system prompt template \(p\) are then combined to form the model input. The LLM \(P\) processes this input to produce both the reasoning trajectory \(\hat{r}_i\) and a predicted answer \(\hat{y}_i\). Each trajectory \(\hat{r}_i\) represents a multi-turn reasoning path, where each step corresponds to a reasoning decision and a tool invocation. During data synthesis, CaptionZoom is the only visual access point: it converts frames from a chosen interval into temporally grounded captions, which the LLM then reasons over. Instead of using raw frames via FrameZoom, the LLM interacts with CaptionZoom, enabling stronger, text-based visual reasoning. To enhance reasoning diversity, we set the LLM’s sampling temperature to 0.7 and generate 5 distinct reasoning trajectories for each input. The LLM iteratively performs reasoning and tool invocations until it reaches a final answer or the predefined step limit \(T\). 
We retain only the trajectories whose predicted answer $\hat{y}_i$ matches the ground-truth $y_i$; if no trajectory satisfies this condition, we randomly select one trajectory from the candidates. Through this process, we obtain diverse, interpretable, and temporally grounded reasoning traces that facilitate robust video reasoning performance.
\begin{algorithm}[t]
\caption{Adaptive Reasoning in VideoThinker}
\label{alg:adaptive_reasoning}
\small
\begin{algorithmic}[1]
\State \textbf{Input:} Video $V$; duration $L$; query $Q$; confidence threshold $\tau$; number of sampled frames $n$; VideoThinker $p_\theta$
\State \textbf{Initialize:} retrieved clips $\mathcal{C} \gets \emptyset$

\If{$L < 600 s $}
    \State Sample frames $\{f_i\}_{i=1}^n \sim \text{Uniform}(V)$
    \State $(\hat{y}, \gamma) \gets p_\theta(\{f_i\}, Q)$ \Comment{Direct reasoning}
\Else
    \State $\mathcal{C} \gets \text{RetrieveTopKClips}(V, Q, k)$
    \State $(\hat{y}, \gamma) \gets p_\theta(\mathcal{C}, Q)$ \Comment{Direct reasoning}
\EndIf

\If{$\gamma < \tau$}
    \State $(\hat{y}, \gamma) \gets p_\theta(V, Q, \mathit{Prompt}_{\text{Tool\_reason}})$ \Comment{Tool reasoning}
\EndIf

\State \Return $\hat{y}$
\end{algorithmic}
\end{algorithm}

\begin{table*}[t]
    \centering
    \renewcommand{\arraystretch}{1.25} 
    \resizebox{\textwidth}{!}{
    \begin{tabular}{l c c c c c}
    \hline
    \rowcolor{gray!12} 
    \textbf{Model} & \textbf{\#Frames} & \textbf{MLVU} & \textbf{LVBench} & \textbf{VideoMME (L)} & \textbf{LongVideoBench} \\ 
    \hline
    \textbf{Closed-source Models} & & & & & \\
    \hline
    GPT-4o~\citep{hurst2024gpt}             & 384 & 54.9  & 48.9         & 72.1  & 66.7  \\ 
    Gemini-1.5-pro~\citep{team2024gemini}    & 256 & --   & 33.1         & 77.4  & 64.0  \\
    Seed1.5VL-pro~\citep{seed2025seed1_5vl} & 32  & 54.9  & 46.1         & 63.3  & 63.7  \\
    \hline
    \textbf{Open-source Models} & & & & & \\
    \hline
    Qwen2.5VL-72B~\citep{bai2025qwen2}         & 128 & 53.8  & 47.4         & 64.6  & 60.3  \\
    LongVILA-7B~\citep{chen2024longvila}     & 256 & 49.0  & --            & 52.1  & 57.7  \\
    Video-XL-7B~\citep{shu2025video}         & 256 & 45.5  & --            & 54.9  & 50.7  \\
    Video-R1-7B~\citep{feng2025video}        & 64  & --   & --            & 52.2  & --     \\
    Qwen2.5VL-7B~\citep{bai2025qwen2}          & 64  & 48.0  & 38.3         & 50.0  & 56.7  \\
    \hline
    \textbf{Agentic LLMs} & & & & & \\
    \hline
    VideoAgent~\citep{wang2024videoagent} (GPT-4)           &   & 52.2  & -  & 46.2  & -   \\
    VideoAgent~\citep{fan2024videoagent} (GPT-4)           &   & 55.4  & -  & 48.1  & -   \\
    VideoTree~\citep{wang2025videotree} (Qwen-plus)         &   & 51.6  & -  & 39.3  & -   \\
    VideoExplorer~\citep{yuan2025thinkvideosagenticlongvideo} (Qwen2.5-7B-tuning + Qwen2.5-VL-32B)  &   & 58.6  & 51.4  & -  & -   \\
    Language Agent (Qwen-235B w/ our tools)         &   & 50.9  & 45.4  & -  & 58.2   \\
    \hline
    \textbf{VideoThinker} (Qwen2.5VL-7B w Think with videos) & 64 &  54.8 (+6.8) &  48.9 (+10.6) &    53.7 (+3.7) &  59.7 (+3.0) \\
    \hline
    \end{tabular}}
\caption{Evaluation results on MLVU, LVBench, VideoMME (L), and LongVideoBench. \textbf{VideoThinker} outperforms open-source models and demonstrates competitive performance with both closed-source models and LLM-based language agents.}
\label{tab:video_results}
\end{table*}

\subsection{Multimodal Tool-Reasoning Training}
\label{method:3}
Based on the tool-augmented reasoning data generated by the LLM, we construct the video-interleaved CoTs dataset
\(\mathcal{D}_{\text{tool}} = \{(v_i, x_i, \hat{r}_i, \hat{y}_i)\}_{i=1}^{M}\).  
Each reasoning trajectory \(\hat{r}_i\) consists of a sequence of reasoning steps and corresponding tool invocations. Among these, the FrameZoom tool replaces the CaptionZoom tool and plays a central role, as it is the only component capable of accessing the video content directly. It retrieves temporally localized visual evidence, providing the LLM with perceptual grounding essential for reasoning. During dataset construction, the textual outputs of Caption\_Zoom are replaced by corresponding video segments represented as special \texttt{<video>} tokens. This conversion transforms the reasoning traces into multimodal interaction sequences, enabling direct visual-textual supervision for the VideoLLM. The multimodal student VideoLLM \(f_{\theta}\) is trained to reproduce both the reasoning process and the final answer by minimizing the combined objective:
\begin{equation}
\label{eq:train_obj}
\mathcal{L} =
\frac{1}{M}\sum_{i=1}^{M}
\Big[
\ell(f_{\theta}^{(1:T)}(v_i, x_i), \hat{r}_i)
\Big],
\end{equation}
where \(\ell(\cdot)\) denotes the token-level cross-entropy loss. Through this training, the student VideoLLM learns to align textual reasoning with visual grounding, internalizing the teacher’s structured tool-usage logic. Consequently, it can perform interpretable, temporally grounded reasoning directly over video frames.

\begin{figure}[t]
\centering
\includegraphics[width=0.8\linewidth]{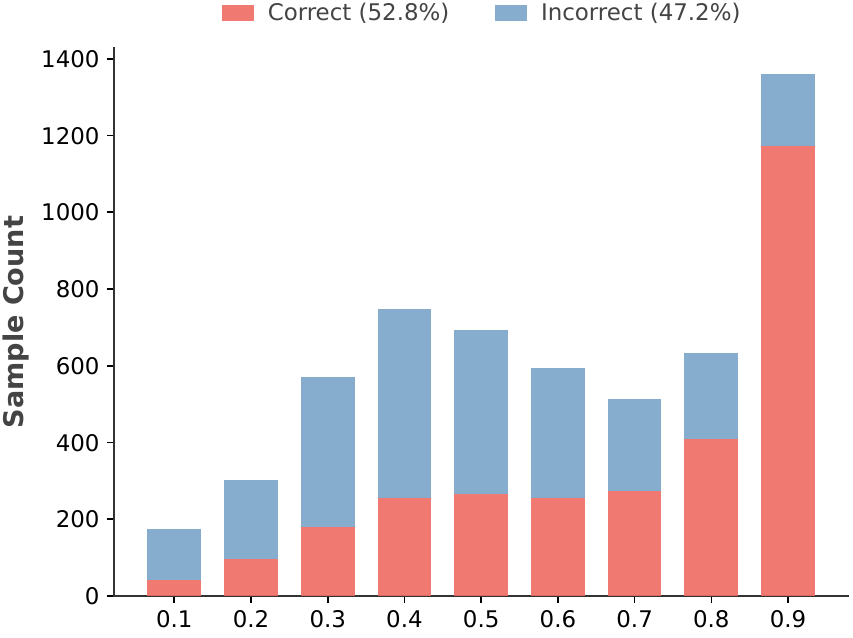}
\caption{Confidence–accuracy relationship. The analysis is conducted on samples from VideoMME (2.7k), LongVideoBench (1.3k), and LVBench (1.5k).
}
\label{fig:confidence}
\end{figure}

\subsection{Adaptive Video Reasoning} 
\label{method:4}
As shown in Figure~\ref{fig:confidence}, model confidence exhibits a strong correlation with prediction accuracy. Approximately 90\% of the samples with confidence scores between 0.9 and 1.0 are correct, whereas more than half of those with confidence below 0.5 are incorrect. Motivated by this observation, we use confidence as a control signal to trigger multi-round tool-based reasoning. Given an input video and a query, VideoThinker first samples $n$ frames if the video duration is shorter than 600 seconds; otherwise, it retrieves $k$ relevant clips. The reasoning model $VideoThinker$ ($p_\theta$) then produces an initial answer along with a confidence score $\gamma$.

\begin{equation}
\gamma = \exp\!\left(\frac{1}{m}\sum_{t=1}^{m}\log f_{\theta}\!\left(\hat{y}_t \mid v, x, \hat{\mathbf{y}}_{<t}\right)\right).
\end{equation}

If $\gamma > \tau$, the answer is directly returned; otherwise, VideoThinker initiates a tool-augmented reasoning process to refine the response using the tool-reasoning prompt. This adaptive two-stage design effectively balances efficiency and accuracy by combining direct reasoning with deeper, tool-guided reasoning under uncertainty.

\section{Experiments}
\subsection{Experimental Setup}
\begin{figure*}[t]
\centering

\begin{minipage}{0.48\linewidth}
    \centering
    \includegraphics[width=1.0\linewidth]{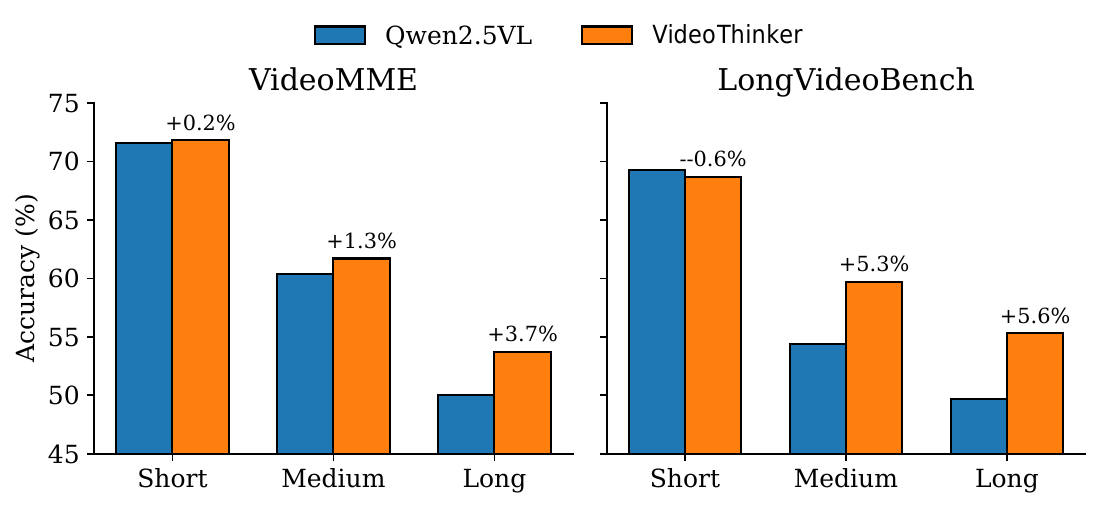}
\caption{Accuracy on LongVideoBench and VideoMME across different video durations. 
Short: $<2$ min; Medium: $2$--$15$ min; Long: $>15$ min.}

    \label{fig:duration_ablation}
\end{minipage}
\hfill
\begin{minipage}{0.48\linewidth}
    \centering
    \includegraphics[width=1.0\linewidth]{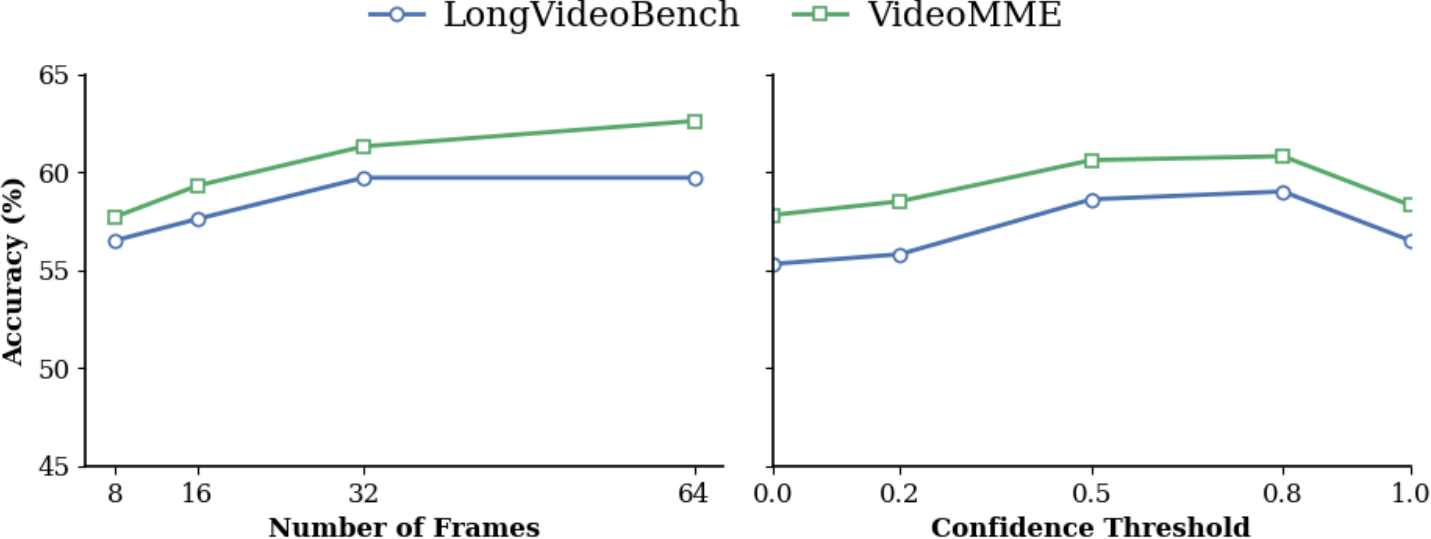}
    \caption{Performance of the VideoThinker at varying confidence threshold on Long-form VideoQA.}
    \label{fig:confid_frames}
\end{minipage}

\end{figure*}
\paragraph{Benchmarks}
To evaluate the performance of our model, we tested it on four distinct long video benchmarks: LongVideoBench \citep{wu2024longvideobench}, VideoMME \citep{fu2024video}, LVBench \citep{wang2024lvbench}, and MLVU~\cite{zhou2025mlvu}. These benchmarks cover a variety of tasks to assess the model’s video comprehension and multimodal reasoning capabilities. LongVideoBench \citep{wu2024longvideobench} is a benchmark for long video understanding, consisting of 3,763 videos (up to 1 hour) with subtitles and 6,678 human-annotated multiple-choice QA pairs across 17 categories. We use the validation dataset for evaluation. VideoMME \citep{fu2024video} contains 900 videos (totaling 254 hours) with 2,700 human-curated QA pairs across six domains and 30 subcategories. The videos range from 11 seconds to 1 hour and include frames, subtitles, and audio, enabling multimodal reasoning. We mainly evaluate using the subset of long videos (over 600 seconds). LVBench \citep{wang2024lvbench} focuses on extreme long video comprehension, with videos ranging from 70 seconds to 4 hours. It includes single-scene, multi-scene, and full-scene settings, covering diverse reasoning types such as temporal, spatial, causal, hypothetical, and external knowledge. MLVU \citep{zhou2025mlvu} is a benchmark for multi-task long video understanding, featuring a diverse set of videos from genres like movies, surveillance, egocentric videos, and cartoons.  We use the test set for evaluation.
\vspace{-1em}
\paragraph{Implementation Details}
We construct the multimodal video-interleaved tool reasoning CoTs based on CG-Bench \citep{chen2024cg}, which includes 10k multiple-choice QA instances. The LLM agent uses Qwen3-235B-A22B-MoE \citep{yang2025qwen3} deployed on 4 NVIDIA H200 GPUs, while the VideoLLM is Qwen2.5-VL-7B \citep{bai2025qwen2}. The VideoLLM is fine-tuned on 10k video-interleaved samples for 3 epochs, utilizing up to 4 NVIDIA H200 GPUs. Training and inference are carried out using MS-Swift \citep{zhao2024swiftascalablelightweightinfrastructure}. For efficient reasoning, VideoThinker uses 32 frames to directly generate the answer and its confidence, with each frame having a maximum resolution of 32,768 pixels. During the two-stage reasoning process, the total number of frames involved does not exceed 64. We evaluate VideoThinker against several baseline models, including closed-source and open-source models:

\begin{itemize}
    \item \textbf{Closed-source models}: GPT-4o \citep{hurst2024gpt}, Gemini-1.5 Pro \citep{team2024gemini}, and Seed 1.5VL-Pro \citep{seed2025seed1_5vl}.
    \item \textbf{Open-source models}: Qwen2.5VL-72B \citep{bai2025qwen2}, LongVILA-7B \citep{chen2024longvila}, Video-XL-7B \citep{shu2025video}, Video-R1-7B \citep{feng2025video}, and Qwen2.5VL-7B \citep{bai2025qwen2}.
    \item \textbf{Agentic LLMs}: VideoAgent (with GPT-4) \citep{wang2024videoagent}, VideoAgent (with GPT-4) \citep{fan2024videoagent}, VideoTree (with Qwen-plus) \citep{wang2025videotree}, and VideoExplorer (Qwen2.5-7B-tuning + Qwen2.5-VL-32B) \citep{yuan2025thinkvideosagenticlongvideo}.
\end{itemize}

For the LVBench and MLVU benchmarks, audio subtitles are generated using FFmpeg and Whisper~\citep{radford2023robust}. All decoding follows the official configurations, with the confidence threshold set to 0.7 by default.


\subsection{Main results}
As shown in Table~\ref{tab:video_results}, VideoThinker consistently outperforms open-source models and remains competitive with both closed-source systems and LLM-based agents. Specifically, VideoThinker achieves 54.8\% on MLVU, surpassing Qwen2.5VL-7B (48.0\%) and Qwen2.5VL-72B (53.8\%), and performing on par with GPT-4o (54.9\%). On LVBench, it reaches 48.9\%, outperforming Qwen2.5VL-7B (38.3\%) and Seed1.5VL-Pro (46.1\%), and matching GPT-4o. For VideoMME, VideoThinker achieves 53.7\%, exceeding Qwen2.5VL-7B (50.0\%). On LongVideoBench, it obtains 59.7\%, surpassing Qwen2.5VL-7B (56.7\%) and closely approaching Qwen2.5VL-72B (60.3\%). VideoExplorer achieves $58.6\%$ on MLVU and $51.4\%$ on LVBench, whereas VideoThinker obtains $54.8\%$ and $48.9\%$ with a single $7$B model. 
In contrast, VideoExplorer is implemented as a language-agent pipeline that couples a separate large-scale LLM with an external Video-LLM, while VideoThinker is a single end-to-end $7$B VideoLLM, highlighting the efficiency of our approach. Notably, in the LLM-based agent setting, we compare against the language agents used in our data synthesis process. 
Compared with this redundant architecture—where VideoLLM acts as the tool, VideoThinker achieves an overall improvement of about 3\%, including a +3.9\% gain on MLVU. Overall, these results highlight VideoThinker's strong performance across benchmarks, showing that our "Thinking with Videos" framework enables competitive long-form video reasoning while maintaining a lightweight design.

\subsection{Ablation studies}
\paragraph{Effect of Video Duration}
As shown in Figure~\ref{fig:duration_ablation}, VideoThinker performs on par with Qwen2.5-VL when handling short videos (under 2 minutes). However, its advantage becomes evident as the video length increases. For medium videos (between 2 and 15 minutes), VideoThinker achieves consistently higher accuracy on both VideoMME and LongVideoBench, showing stronger capability in modeling longer temporal dependencies. 
The improvement is even more significant on long videos (exceeding 15 minutes), where VideoThinker surpasses Qwen2.5-VL by 3.7 percentage points on VideoMME and 5.6 percentage points on LongVideoBench. These results demonstrate that VideoThinker maintains stable reasoning and robust understanding in extended temporal contexts, effectively addressing the information loss and temporal drift commonly encountered in long-form video understanding.

\paragraph{Impact of $n$ and $\tau$ on Performance}
We analyze the effects of the number of frames ($n$) and the confidence threshold ($\tau$) on model performance. 
As shown in Figure~\ref{fig:confid_frames}, increasing $n$ from 8 to 64 significantly improves accuracy—from 56.5\% to 59.7\% on LongVideoBench and from 57.7\% to 62.6\% on VideoMME—indicating that denser sampling provides richer visual context for reasoning. 
For the confidence threshold $\tau$, which controls the switch between fast and tool-based reasoning, performance improves as $\tau$ increases from 0.0 to 0.8 (from 55.3\% to 59.0\% on LongVideoBench and from 57.8\% to 60.8\% on VideoMME), but drops when $\tau=1.0$ (56.5\% and 58.3\%, respectively). 
A low $\tau$ causes overreliance on direct reasoning, while a high $\tau$ triggers excessive tool reasoning. These results suggest that adaptive confidence-gated reasoning, combined with more frame samples, yields the best overall performance.

\begin{table}[t]
\centering
\scriptsize
\setlength{\tabcolsep}{4.5pt}
\begin{tabular}{lcccccccc}
\toprule
\multirow{2}{*}{\textbf{Dataset}} 
& \multicolumn{4}{c}{\textbf{Retrieval\_clips (Top-$k$)}} 
& \multicolumn{4}{c}{\textbf{Retrieval\_subtitles (Top-$k$)}} \\
\cmidrule(lr){2-5} \cmidrule(lr){6-9}
& 1 & 3 & 5 & 10 & 1 & 3 & 5 & 10 \\
\midrule

\textbf{LongVideoBench} 
& 59.7 & \textbf{59.8} & 58.8 & 58.6 
& 57.4 & \textbf{58.4} & 58.0 & 57.1 \\

\textbf{VideoMME (L)}
& 53.7 & 52.3 & 54.2 & \textbf{54.7} 
& 51.2 & 50.2 & \textbf{53.1} & 52.3 \\

\textbf{LVBench} 
& 48.9 & \textbf{49.1} & 47.2 & 49.0 
& \textbf{50.6} & 50.5 & 49.0 & 49.3 \\

\bottomrule
\end{tabular}
\caption{Ablation of \textbf{Retrieval\_clips} and \textbf{Retrieval\_subtitles} with varying Top-$k$ values ($1,3,5,10$) on three datasets. }
\label{tab:retrieval_topk_singlecol}
\end{table}

\begin{figure}[t]
\centering
\includegraphics[width=0.8\linewidth]{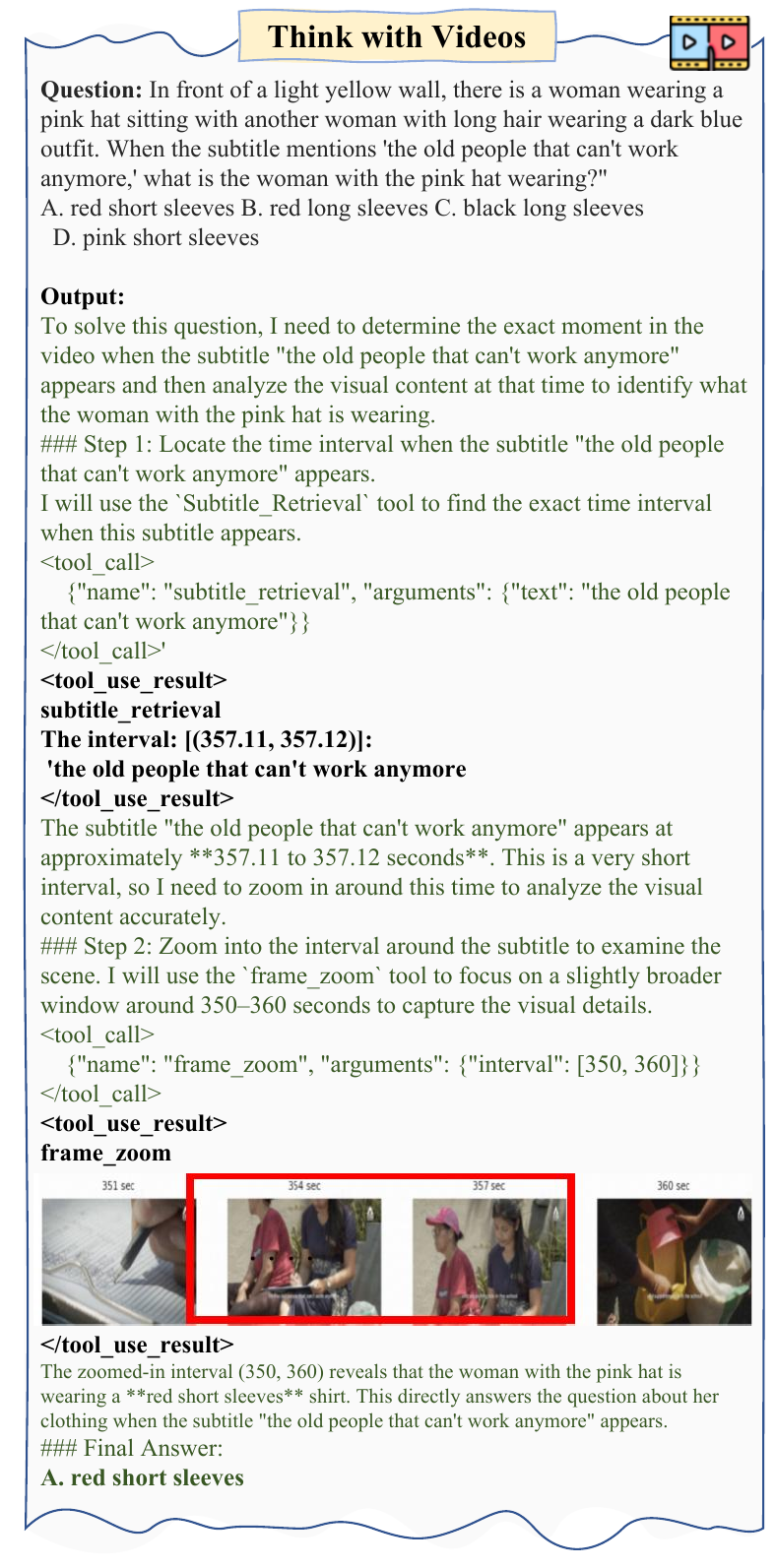}
\caption{Case Study: VideoThinker performs agentic tool use by retrieving subtitles to locate temporal intervals and zooming into relevant frames for video reasoning.}
\label{fig:case}
\end{figure}

\paragraph{Ablation on Retrieval Tools}
We analyze the effect of retrieval by varying the number of retrieved clips and subtitles (\textit{Top-}$k$). As shown in Table~\ref{tab:retrieval_topk_singlecol}, \textit{Top-}1 is generally suboptimal, indicating that a single retrieved item provides limited context. Increasing $k$ typically improves performance as the model benefits from richer and more diverse information. On \textbf{LongVideoBench}, the best results appear at \textit{Top-}3 for clips (59.8) and subtitles (58.4), showing a good balance between relevance and noise. \textbf{VideoMME (L)} favors broader context, peaking at \textit{Top-}10 for clips (54.7) and \textit{Top-}5 for subtitles (53.1). Top-1 is often slightly worse or comparable to larger k, though on LVBench subtitle retrieval, \textit{Top-}-1 performs best, indicating that the optimal k can be dataset-dependent.


\subsection{Case Study}
\label{sec:case}

To further illustrate the multi-step reasoning process of VideoThinker, we present a representative case in Figure~\ref{fig:case}. The model is asked the question:  \textit{``When the subtitle mentions ‘the old people that can’t work anymore,’ what is the woman with the pink hat wearing?''}  
VideoThinker first invokes the Subtitle\_Retrieval tool to locate the temporal position of the mentioned subtitle. The system accurately identifies the corresponding timestamp at 357.11s, which serves as a temporal anchor for subsequent reasoning. Based on the retrieved timestamp, VideoThinker employs the frame\_zoom tools to examine the visual context within a narrow temporal window (350–360s). It focuses on frames depicting the woman with the pink hat and analyzes her appearance in detail. Through this fine-grained inspection, the model correctly determines that the woman is wearing a red short-sleeve shirt.

\section{Conclusion}

In this work, we present VideoThinker, an agentic VideoLLM designed for long-form video understanding through adaptive Temporal Retrieval and Temporal Zoom tool use. By leveraging temporal retrieval and temporal zoom, VideoThinker enables adaptive exploration of key video moments, allowing for dynamic reasoning over long videos. Trained on synthetic tool-interaction data generated through LLM-guided reasoning, VideoThinker learns to reason across long videos without requiring pre-existing long-form understanding. Extensive evaluations on multiple long-form video benchmarks demonstrate that VideoThinker consistently outperforms strong VideoLLM and LLM-agent baselines, validating the effectiveness of adaptive, tool-augmented reasoning for long-form video understanding.

\*section\section*{Acknowledgments}

This work was supported by the NSFC project (No.~62072399), the Zhejiang Provincial Natural Science Foundation of China under Grant No. LZ23F020009, the Fundamental Research Funds for the Central Universities (No.~S20240030), MoE Engineering Research Center of Digital Library, China Research Centre on Data and Knowledge for Engineering Sciences and Technology.
\clearpage
\setcounter{page}{1}
\maketitlesupplementary
\appendix
\section{Supplementary Materials}
\subsection{Prompt Details}
\label{appendix:prompt}

\begin{figure}[H]
    \centering \includegraphics[width=0.9\linewidth]{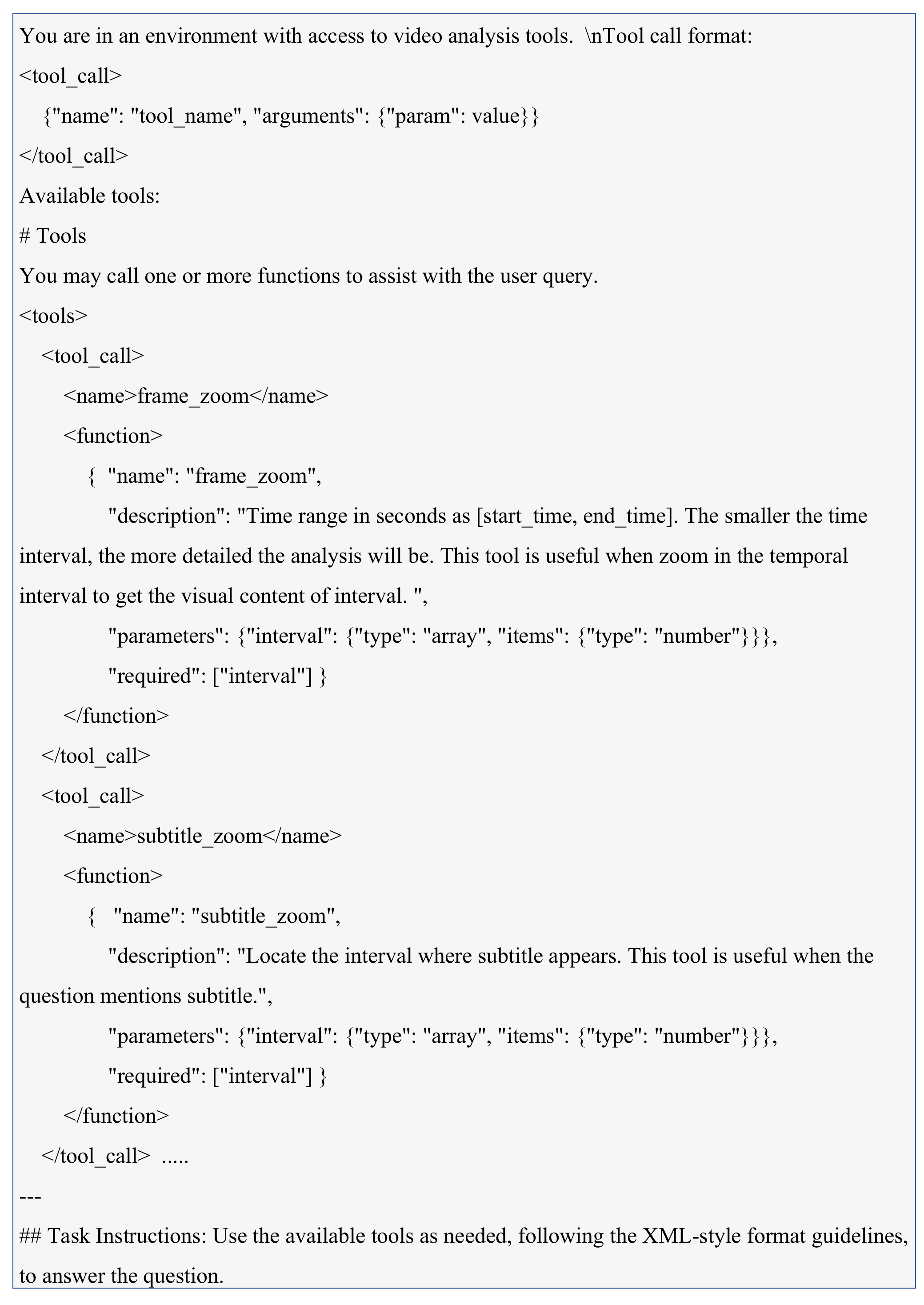}
    \caption{The prompt is designed to enable VideoLLM to think with videos using the available tools.}
    \label{figure:tool_prompt}
\end{figure}

As shown in Figure~\ref{figure:tool_prompt}, we provide the detailed Tool reasoning prompt used in our framework. 
This prompt specifies the expected output format for tool calls, defines all available tools, and outlines the instructions for each task. For comparison, the Direct reasoning prompt is defined as follows:
\begin{quote}
Question: \{question\}, output a single letter. Best option:
\end{quote}
\subsection{Tools }
As shown in Table~\ref{tab:tool_suite}, our framework provides a set of tools for long-form video reasoning at multiple levels of granularity. 
The toolkit combines retrieval modules, such as Clip Retrieval and Subtitle Retrieval, with zoom-level tools, including Frame Zoom, Subtitle Zoom, and Caption Zoom, to support both coarse- and fine-grained analysis. 
The Subtitle Summary module further offers high-level textual abstraction. 
Together, these tools enable effective and interpretable reasoning over long-duration videos. 

\begin{figure}
    \centering
    \includegraphics[width=0.9\linewidth]{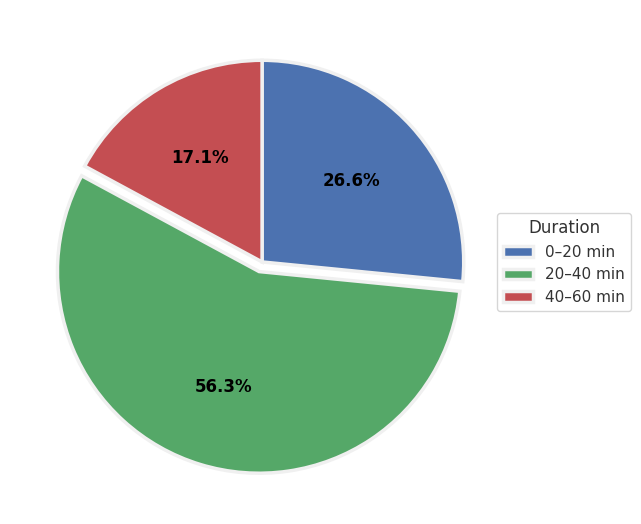}
    \caption{Distribution of Video Durations in CoTs.}
    \label{fig:sft_duratuon}
\end{figure}

\subsection{Training Data}
We analyze the distribution of video lengths in the CG-bench synthetic dataset, as illustrated in Figure~\ref{fig:sft_duratuon}. 
More than half of the videos have durations between 20 and 40 minutes, while those lasting 40- 60 minutes account for 17.1\%. Most data samples involve approximately 3-5 tool calls.

\begin{table*}[ht]
\centering
\resizebox{\textwidth}{!}{ 
\begin{tabular}{|>{\centering\arraybackslash}m{0.18\textwidth}| 
                >{\raggedright\arraybackslash}m{0.25\textwidth}| 
                >{\raggedright\arraybackslash}m{0.52\textwidth}|}
\hline
\textbf{Tool} & \textbf{Parameters} & \textbf{Description} \\
\hline

Clip\_Retrieval & 
video path, query, topk &
Retrieves the top-$k$ video clips most relevant to the given query based on semantic similarity. \\

\hline
Subtitle\_Retrieval & 
video path, query, topk &
Retrieves the top-$k$ subtitle segments relevant to the query from Whisper-transcribed subtitles based on semantic similarity. \\

\hline
Subtitle\_Summary &
video path, query &
Summarizes subtitles using an LLM to provide concise, query-focused contextual understanding for reasoning. \\

\hline
Frame\_Zoom & 
video path, interval &
Extracts video frames from the specified time interval. \\

\hline
Subtitle\_Zoom & 
video path, interval &
Extracts subtitles from the specified time interval. \\

\hline
Caption\_Zoom & 
video path, interval &
Extracts visual captions from the specified time interval. \\

\hline
\end{tabular}
}
\caption{Tool suites for long-form video reasoning, integrating retrieval-based and zoom-level analysis capabilities.}
\label{tab:tool_suite}
\end{table*}
\begin{table*}[htbp]
\centering
\setlength{\tabcolsep}{10pt}
\begin{tabular}{lccccccc}
\toprule
Methods & ER (\%) & EU (\%) & KIR (\%) & TG (\%) & Rea (\%) & Sum (\%) & Overall (\%) \\
\midrule
\multicolumn{8}{l}{\textit{VideoLLMs}} \\
\midrule
Gemini-1.5-Pro \citep{team2024gemini} & 32.1 & 30.9 & 39.3 & 31.8 & 27.0 & 32.8 & 33.1 \\
GPT-4o \citep{hurst2024gpt}           & 48.9 & 49.5 & 48.1 & 40.9 & 50.3 & 50.0 & 48.9 \\
Qwen2.5-VL-72B \citep{bai2025qwen2}     & --   & --   & --   & --   & --   & --   & 47.4 \\
VideoChat-Flash \citep{li2024videochat} & 51.1 & 46.0 & 49.0 & 38.9 & 48.5 & 34.5 & 48.2 \\
\midrule
\multicolumn{8}{l}{\textit{Agentic LLMs}} \\
\midrule
VideoTree \citep{wang2025videotree}    & 30.3 & 25.1 & 26.5 & 27.7 & 31.9 & 25.5 & 28.8 \\
VideoAgent \citep{wang2024videoagent}  & 28.0 & 30.3 & 28.0 & 29.3 & 28.0 & 36.4 & 29.3 \\
\midrule
\textbf{VideoThinker (Ours)}               & 49.6 & 48.5 & 58.0 & 43.6 & 45.5 & 35.2 & \textbf{48.9} \\
\bottomrule
\end{tabular}
\caption{Comparison on LVBench across various tasks: Entity Recognition (ER), Event Understanding (EU), Key Information Retrieval (KIR), Temporal Grounding (TG), Reasoning (Rea), Summarization (Sum), and Overall performance.}
\label{tab:lvbench}
\end{table*}

\begin{figure}
    \centering \includegraphics[width=0.9\linewidth]{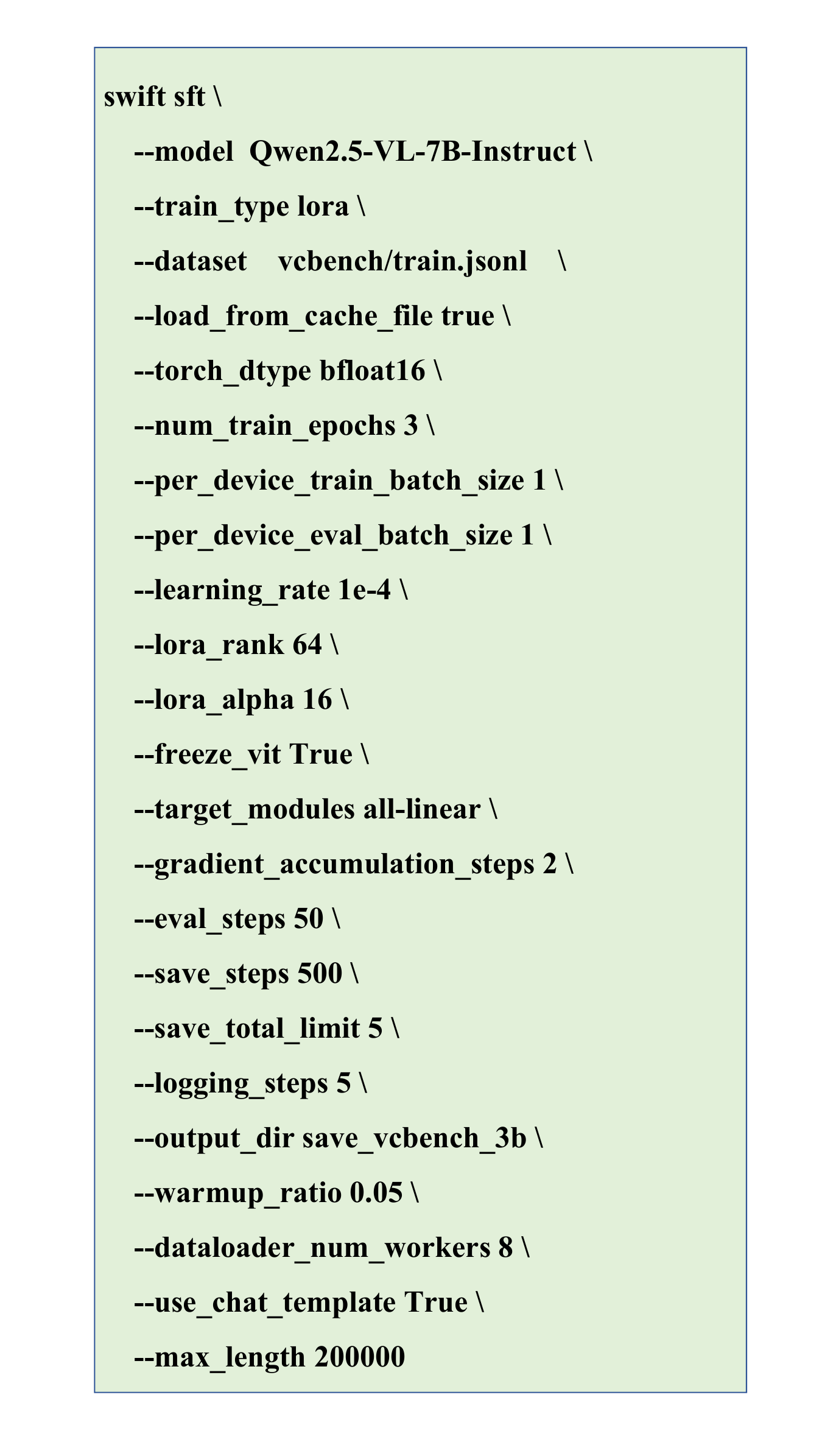}
    \caption{The training script with Swift.}
    \label{figure:training}
\end{figure}

\subsection{Implementation Details}

We present the training script implemented with Swift, as shown in Figure~\ref{figure:training}.

\subsection{More results}
\begin{figure*}[t]
\centering
\includegraphics[width=1.0\linewidth]{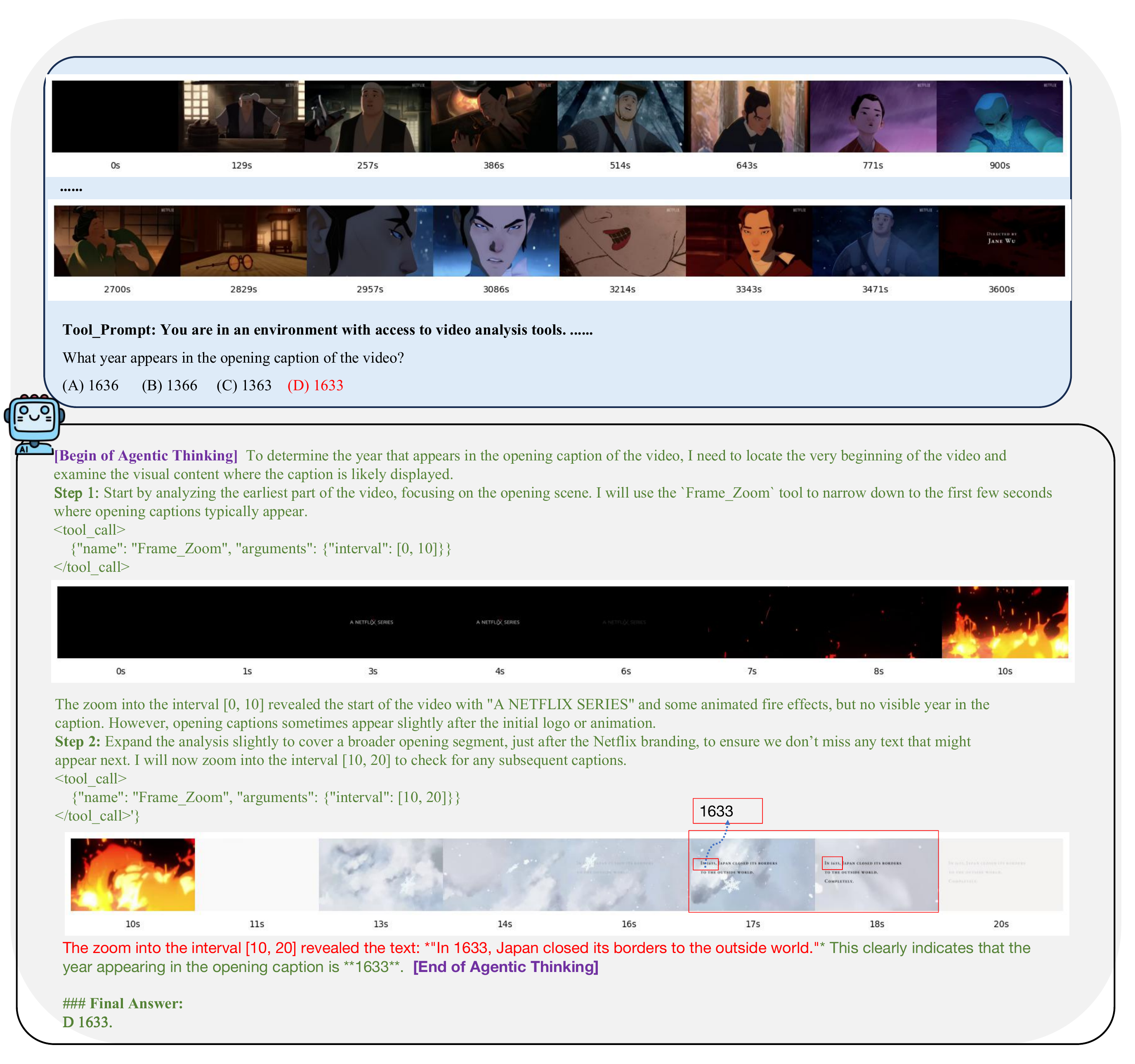}
\caption{VideoThinker’s agentic tool reasoning on LVBench, testing its ability to retrieve key information from a 61-minute video (Cm73ma6Ibcs).}
\label{fig:case1}
\end{figure*}

\begin{figure*}[t]
\centering
\includegraphics[width=0.9\linewidth]{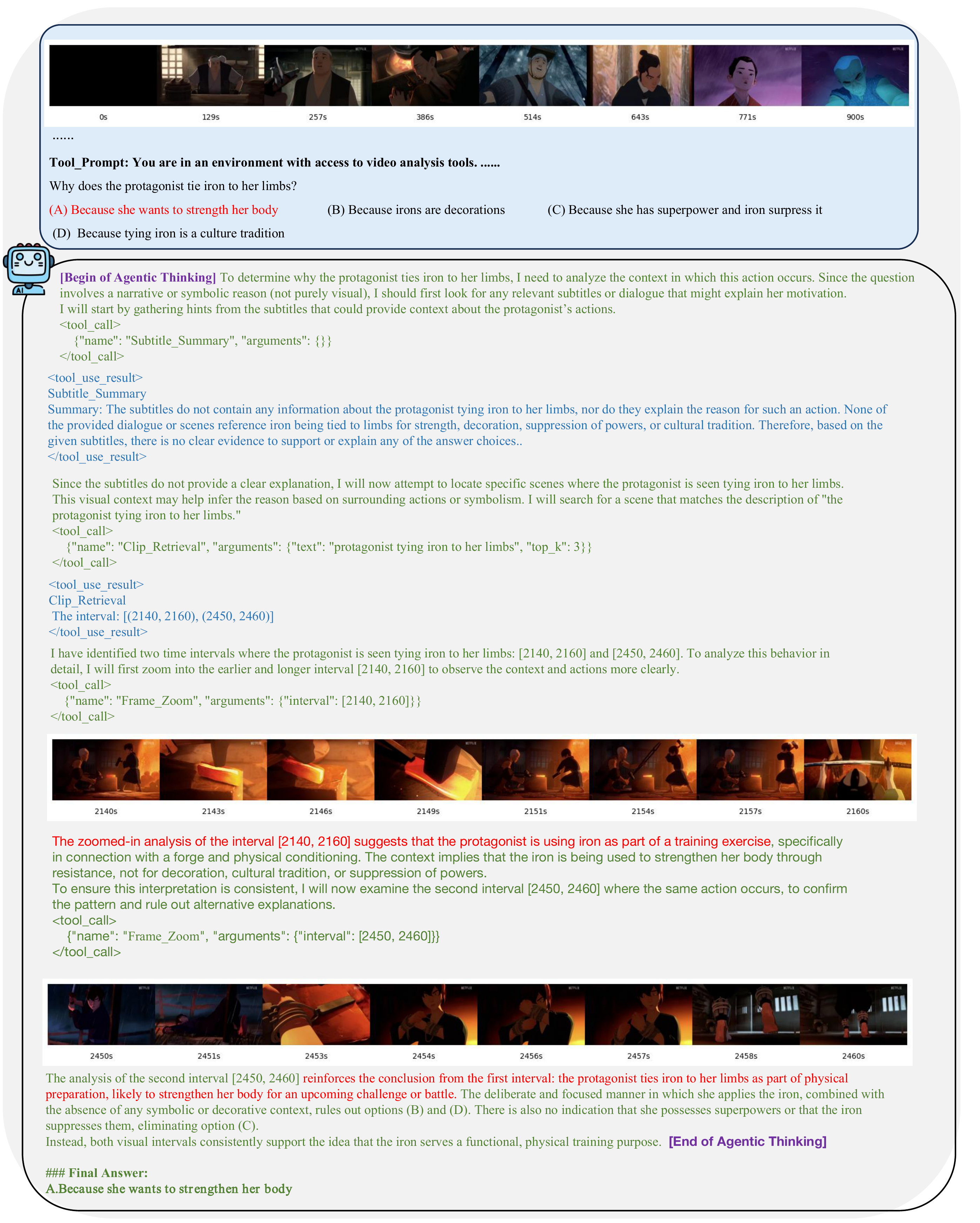}
\caption{VideoThinker’s agentic tool reasoning on LVBench, testing its reasoning ability from a 61-minute video (Cm73ma6Ibcs).}
\label{fig:case2}
\end{figure*}

\begin{figure*}[t]
\centering
\includegraphics[width=1.0\linewidth]{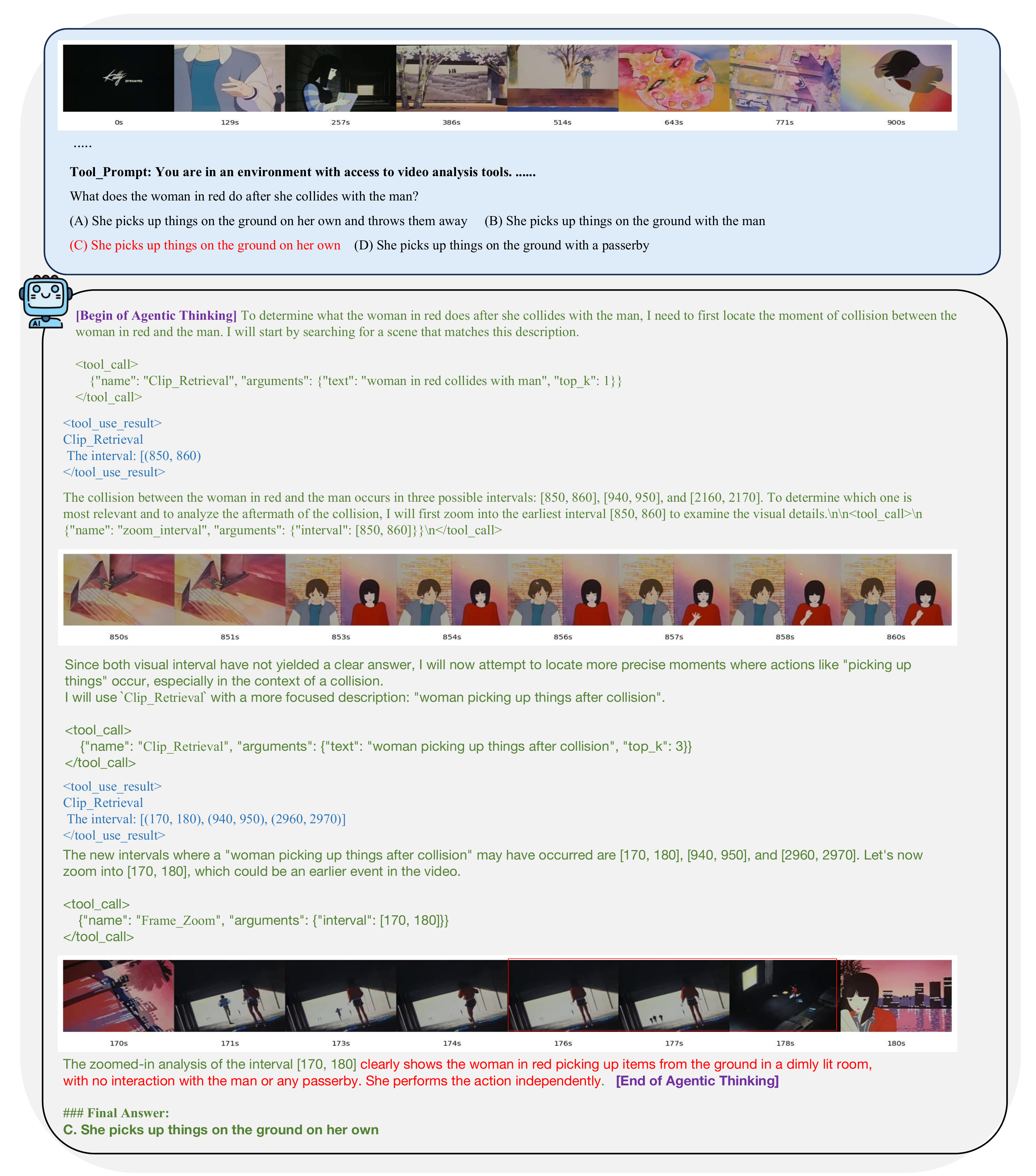}
\caption{VideoThinker’s agentic tool reasoning on LVBench, testing its event-understanding ability using a 55-minute video (q01CUy\_gwdU).}
\label{fig:case3}
\end{figure*}

\begin{figure*}[t]
\centering
\includegraphics[width=0.9\linewidth]{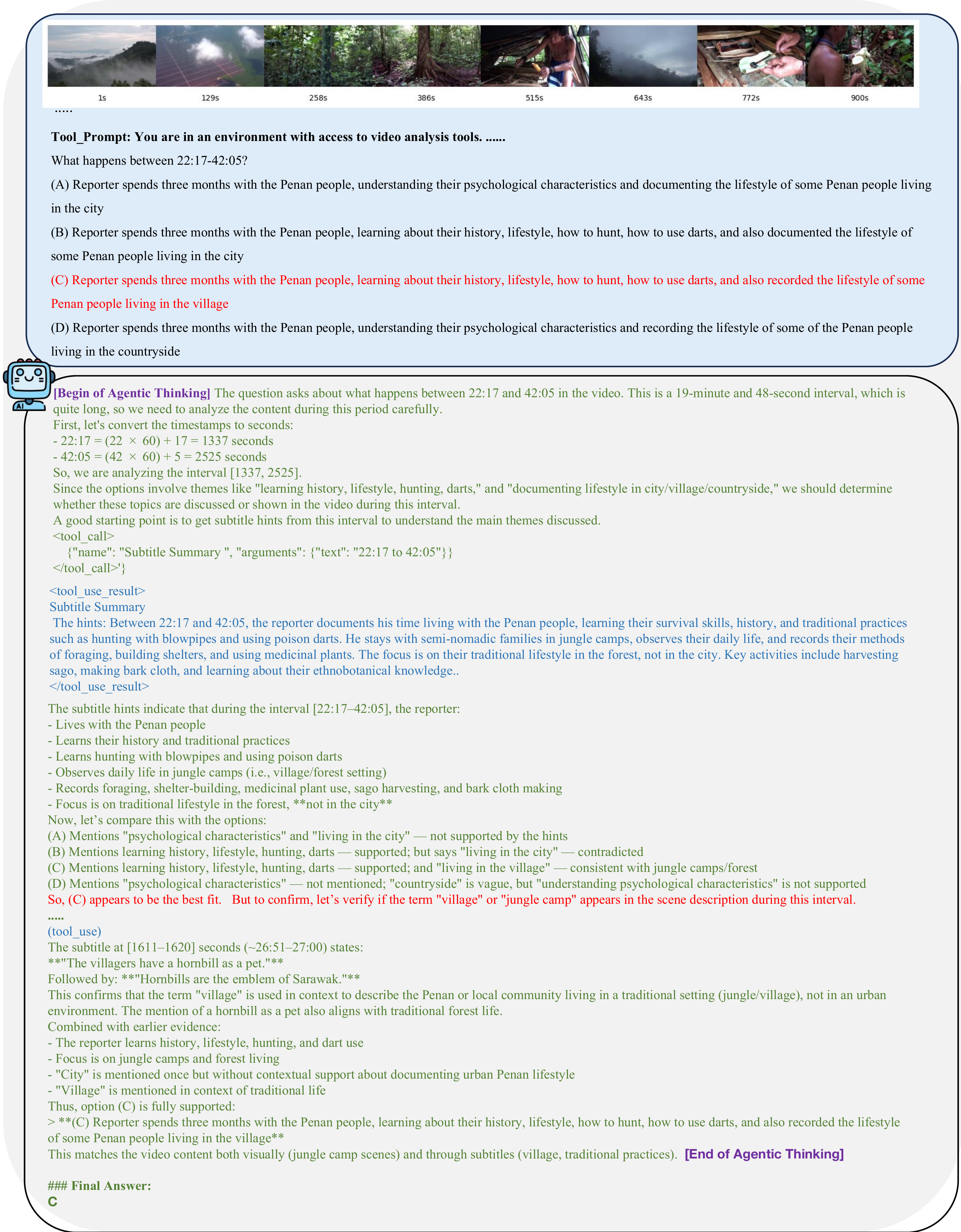}
\caption{VideoThinker’s agentic tool reasoning on LVBench, testing its summarization ability using a 53-minute video (TiQBTesZUJQ).}
\label{fig:case6}
\end{figure*}

Table~\ref{tab:lvbench} presents the comparison results on LVBench across six tasks, including Entity Recognition, Event Understanding, Key Information Retrieval, Temporal Grounding, Reasoning, and Summarization. 
Our method VideoThinker achieves the best overall performance with an average accuracy of 48.9\%, comparable to GPT-4o (48.9\%) and clearly outperforming previous LLM-based agents such as VideoTree (28.8\%) and VideoAgent (29.3\%). 
ReZoom-V also shows strong results on KIR (58.0\%) and ER (49.6\%), demonstrating its effectiveness in retrieval tasks. We present additional LVBench cases in Figure~\ref{fig:case1}--\ref{fig:case6}.

\begin{figure*}[t]
\centering
\includegraphics[width=1.0\linewidth]{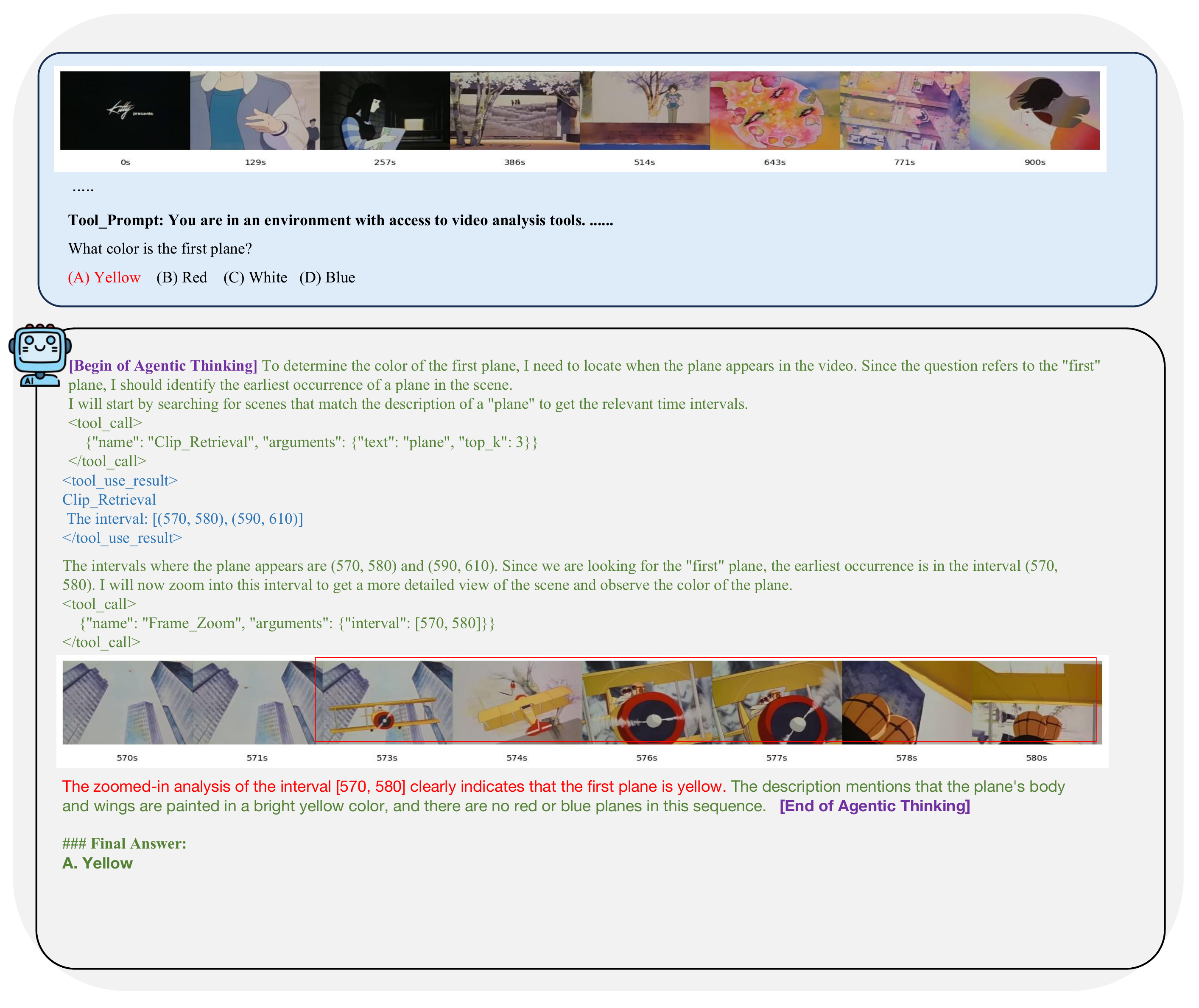}
\caption{VideoThinker’s agentic tool reasoning on LVBench, testing its event-recognition ability using a 55-minute video (q01CUy\_gwdU).}
\label{fig:case4}
\end{figure*}

\begin{figure*}[t]
\centering
\includegraphics[width=1.0\linewidth]{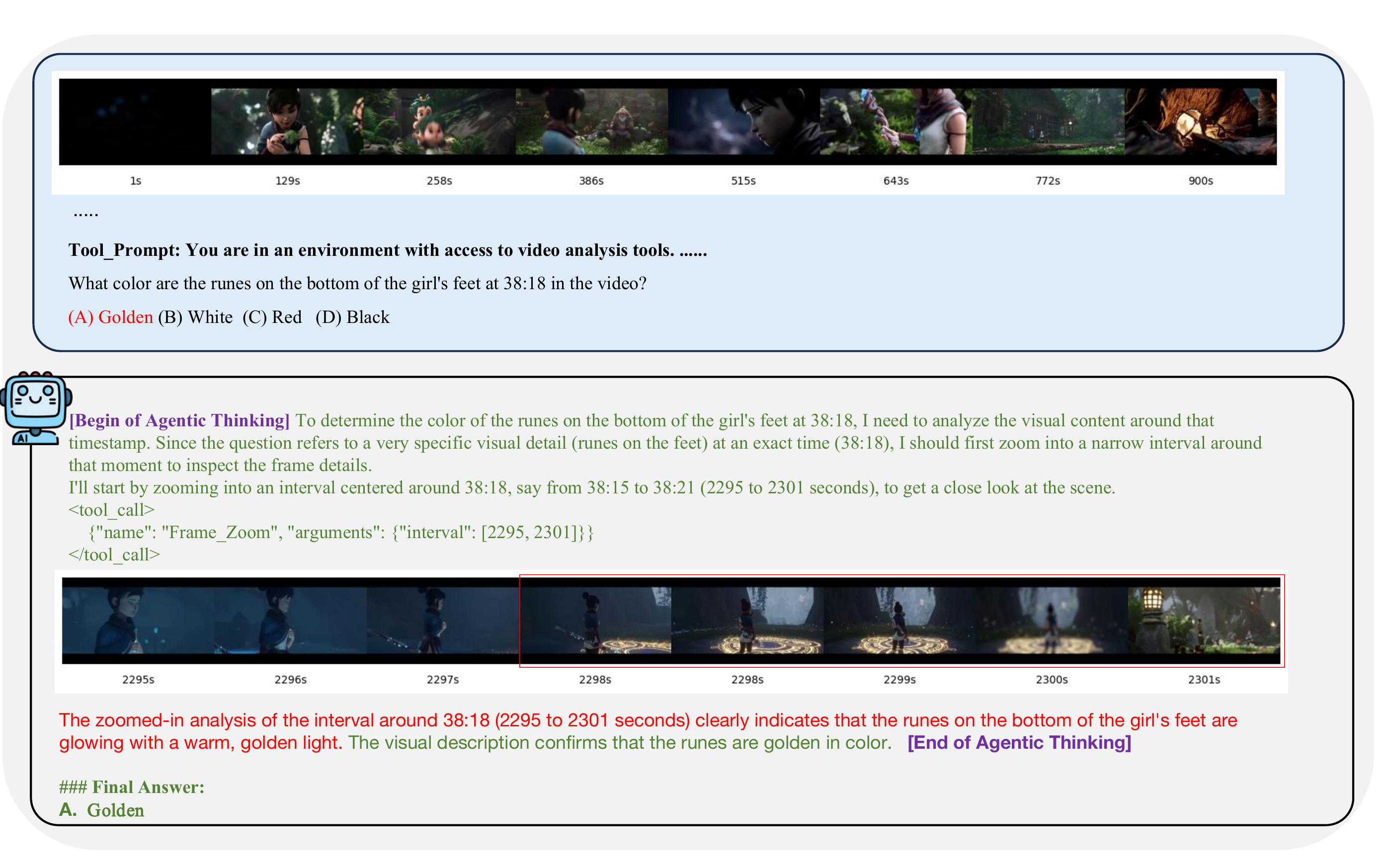}
\caption{VideoThinker’s agentic tool reasoning on LVBench, testing its temporal grounding ability using a 64-minute video (28CIeC8cZks).}
\label{fig:case4}
\end{figure*}

\clearpage

{
    \small
    \bibliographystyle{ieeenat_fullname}
    \bibliography{main}
}

\end{document}